\documentclass[11pt]{article}

\usepackage[preprint]{acl}

\usepackage{times}
\usepackage{latexsym}
\usepackage{xeCJK}

\usepackage[T1]{fontenc}
\usepackage[utf8]{inputenc}

\usepackage{microtype}

\usepackage{inconsolata}

\usepackage{graphicx}

\usepackage{amsmath}
\usepackage{amsfonts}
\usepackage{amssymb}
\usepackage{subcaption}
\usepackage{caption}
\usepackage{graphicx}
\usepackage{booktabs}
\usepackage{multirow}
\usepackage{comment}
\usepackage[capitalize]{cleveref}

\title{LLM NL2SQL Robustness: Surface Noise vs. Linguistic Variation in Traditional and Agentic Settings}

\author{Lifu Tu, Rongguang Wang, Tao Sheng, Sujjith Ravi, Dan Roth \\ \\
 \textbf{Oracle AI} \\}

\begin{document}
\maketitle
\begin{abstract}
Robustness evaluation for Natural Language to SQL (NL2SQL) systems is essential because real-world database environments are dynamic, noisy, and continuously evolving, whereas conventional benchmark evaluations typically assume static schemas and well-formed user inputs. In this work, we introduce a robustness evaluation benchmark containing approximately ten types of perturbations and conduct evaluations under both traditional and agentic settings. We assess multiple state-of-the-art large language models (LLMs), including Grok-4.1, Gemini-3-Pro, Claude-Opus-4.6, and GPT-5.2. Our results show that these models generally maintain strong performance under several perturbations; however, notable performance degradation is observed for surface-level noise (e.g., character-level corruption) and linguistic variation that preserves semantics while altering lexical or syntactic forms. Furthermore, we observe that surface-level noise causes larger performance drops in traditional pipelines, whereas linguistic variation presents greater challenges in agentic settings. These findings highlight the remaining challenges in achieving robust NL2SQL systems, particularly in handling linguistic variability.

\end{abstract}

\section{Introduction}

Natural Language to SQL (NL2SQL) systems aim to translate natural language questions into executable SQL queries, enabling users to interact with databases without requiring knowledge of database query languages. With the rapid progress of large language models (LLMs), recent NL2SQL systems~\citep{dong2023c3zeroshottexttosqlchatgpt,pourreza2023dinsql,dail_sql} have achieved strong performance on widely used benchmarks such as Spider~\citep{yu-etal-2018-spider}. However, these benchmarks typically assume static database schemas and well-formed natural language queries, which differ significantly from real-world deployment scenarios. In practice, database environments are dynamic, user queries may contain noise or ambiguity, and linguistic expressions can vary substantially while preserving the same semantic intent. Existing evaluations\citep{chang2023dr,safarzadeh-etal-2025-evaluating} show that NL2SQL systems remain sensitive to small perturbations. However, prior work largely focuses on single-pass query generation pipelines and does not fully examine how modern agentic NL2SQL systems\citep{lei2024spider}—which incorporate iterative reasoning, tool use, and execution feedback—behave under such perturbations. Consequently, the robustness of both traditional and agent-based NL2SQL systems remains insufficiently understood.

\begin{table*}[ht]
\footnotesize
\centering

\setlength{\tabcolsep}{1pt}
\scalebox{0.70}{
\begin{tabular}{|c|c|} \toprule
Perturbations & \multicolumn{1}{|c|}{Query}                                                                             \\ \toprule
origin                      & In which year were the two most common causes of traffic accidents different from those in other years?          \\
\midrule
BackTranslation (bt)              & In which year were the two most common causes of traffic accidents different from other years?                       \\
Chinese (zh)             &    在哪一年，交通事故最常见的两个原因与其他年份不同?                 \\
Past      &   In which year \textbf{was} the two most common causes of traffic accidents different from those in other years?      \\
Future    & In which year \textbf{will be} the two most common causes of traffic accidents different from those in other years?    \\
InflectionalVariation & In which year \textbf{been} the two most commoner causes of traffic accident different from those in other year?  \\
\midrule
ButterFingers    & In which year w\textbf{t}re th\textbf{v} two mos\textbf{b} \textbf{s}ommo\textbf{u} causes of tr\textbf{w}ffic accidents different from those in other years?          \\
ChangeCharCase               & in w\textbf{H}ich y\textbf{EA}r we\textbf{R}e t\textbf{H}e \textbf{tWO} mos\textbf{T} c\textbf{OM}mo\textbf{N} \textbf{CA}u\textbf{SE}s of \textbf{T}ra\textbf{FF}ic acci\textbf{D}ents d\textbf{IF}fe\textbf{R}ent from tho\textbf{SE} in ot\textbf{H}er yea\textbf{R}s?          \\
%InflectionalVariation & In which year \textbf{been} the two most commoner causes of traffic accident different from those in other year?  \\
SwapCharacters   & In which \textbf{ey}ar we\textbf{er} the two most common causes of traffic accidents different \textbf{rf}om those in \textbf{toeh}r years? \\
Whitespace       & In \textbf{whichy earwere t h e twomost comm on} causes of \textbf{trafficaccidents} \textbf{diffe r ent} from those in \textbf{o ther} years?
\\ \bottomrule
\end{tabular}
}
\caption{Perturbations of a user query: the last two blocks present linguistic variation (preserves semantics while altering lexical or syntactic forms) and surface-level noise (e.g., character-level corruption).}
\label{tab:example}

\end{table*}

To address this gap, we introduce a robustness evaluation framework that systematically examines NL2SQL performance under a diverse set of perturbations. Specifically, we construct approximately ten types of perturbations that simulate realistic challenges, including surface-level noise such as character-level corruption and linguistic variations that preserve semantic meaning while altering lexical or syntactic forms. Using this benchmark, we conduct a comprehensive empirical evaluation across several state-of-the-art large language models, including Grok-4.1, Gemini-3-Pro, Claude Opus 4.6, and GPT-5.2. We evaluate these models under both traditional single-pass generation settings and agentic settings that involve multi-step reasoning and tool interaction.

Our study reveals that while modern LLM-based NL2SQL systems maintain strong performance under certain perturbations, they exhibit notable robustness weaknesses under others. In particular, surface-level noise leads to larger performance degradation in traditional pipelines, whereas linguistic variations pose greater challenges in agentic settings. %These findings highlight the need for improved robustness evaluation and system design for NL2SQL systems operating in realistic environments.
By providing a systematic perturbation-based benchmark and comparative analysis across different system paradigms, our work contributes toward a deeper understanding of robustness challenges in modern NL2SQL systems.

\section{Setup}

\subsection{Text2SQL Task}

Text-to-SQL is the task of automatically translating a natural language question into a structured SQL query that can be executed on a relational database to retrieve the correct answer. Formally, given a database schema $D$ and auxiliary documentation $\epsilon$, a Text-to-SQL system $f$ maps a user query $Q$ to a corresponding SQL query $s$, such that $s$ can be executed over $D$ to produce the intended result:
\begin{equation}
    s = f(q, D, \epsilon) \nonumber
\end{equation}

\paragraph{Traditional Setting.} In this setting, the process typically consists of two stages. The first stage is schema linking, which identifies the tables and columns relevant to the user query. The retrieved schema elements are then provided as additional context alongside the original query. In the second stage, the final SQL statement is generated based on the combined information from the query and the linked schema. Generally, there is no interaction with the database during generation, and no iterative refinement is performed.% In this setting, there are usually two steps. The first step is schema linking, which is to retrieve the relevance tables and columns given the query. With the schema linking results, the linked tables columns are put as an addition context with the query. In the second step, the final SQL is generated based on all the info from the previous step. Usually, there is no interaction with database for SQL refinements. 

\paragraph{Agentic Setting.} Inspired by real-world text-to-SQL workflows, the agentic setting has recently been adopted to generate final SQL through multiple interactions with the database. In this paradigm, the model iteratively refines queries by leveraging execution feedback, debugging errors, and actively exploring the database schema. For complex tasks—such as those in Spider 2.0~\citep{lei2024spider}, agent-based frameworks have demonstrated more promising results compared to traditional state-of-the-art approaches.

\subsection{Transformations}
\label{sec:transform}

To evaluate the robustness of the Text-to-SQL model $f(q, D, \epsilon)$, this work considers two types of variations: (1) perturbations applied to the user query $q$, and (2) updates made to the database $D$.

\paragraph{Query Transformation.}

\begin{figure*}[ht]
  \centering
  \includegraphics[width=0.75\textwidth]{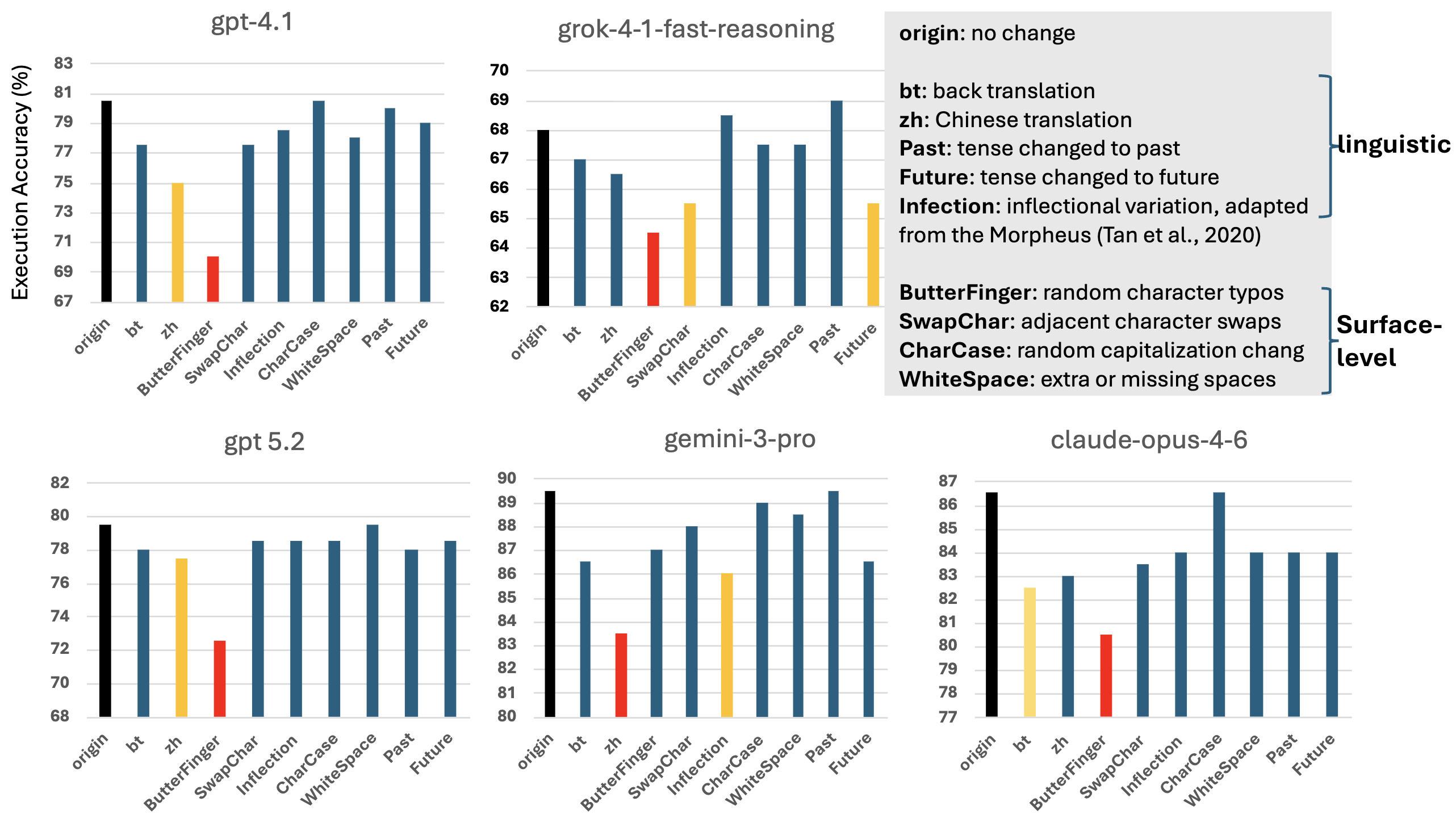}
  \caption{Results on \textbf{R-NL2SQL-Tradition} (adapt from Spider 1~\citep{yu-etal-2018-spider}). LLMs generally maintain strong performance under several perturbations; however, notable performance degradation is still observed.}
  \label{fig:spider1}
\end{figure*}

A user query can vary greatly when written by different users, robustness against changes in query question is critical for usability in applications. Inspired by the NL-Augmenter~\citep{dhole2022nlaugmenter}, which is designed for data augmentation and robustness evaluation, we use two different types noises (surface-level noise and linguistic variation).  %we use the following ten transformations: ``Origin'', ``BackTranslation'', ``Chinese'', ``ButterFinger'', ``ChangeCharCase'' ``InflectionalVariation'', ``SwapCharacter'', ``Past'', ``Future'', ``Whitespace''. ``Origin'' is no change for the original query.  ``BackTranslation'' is the paraphrase from back translation. "Chinese" is the Chinese Translation from English query. `ButterFinger'' introduces random character typos. ``ChangeCharCase'' applied random capitalization changes)
 %``EnglishInflection'' modified singular/plural noun or verb tense variation. ``SwapChar'' swaps adjacent character. ``Future'' shifts the tense to future form.``Past'' shifts tense to past form.``WhiteSpace'' inserts or removes extra spaces. 
 Table~\ref{tab:example} illustrates the perturbation on a sampled user query. More details are shown in Figure~\ref{fig:spider1} and appendix~\ref{sec:appendixA}.

\begin{figure*}[!ht]
  \centering
  \includegraphics[width=0.8\textwidth]{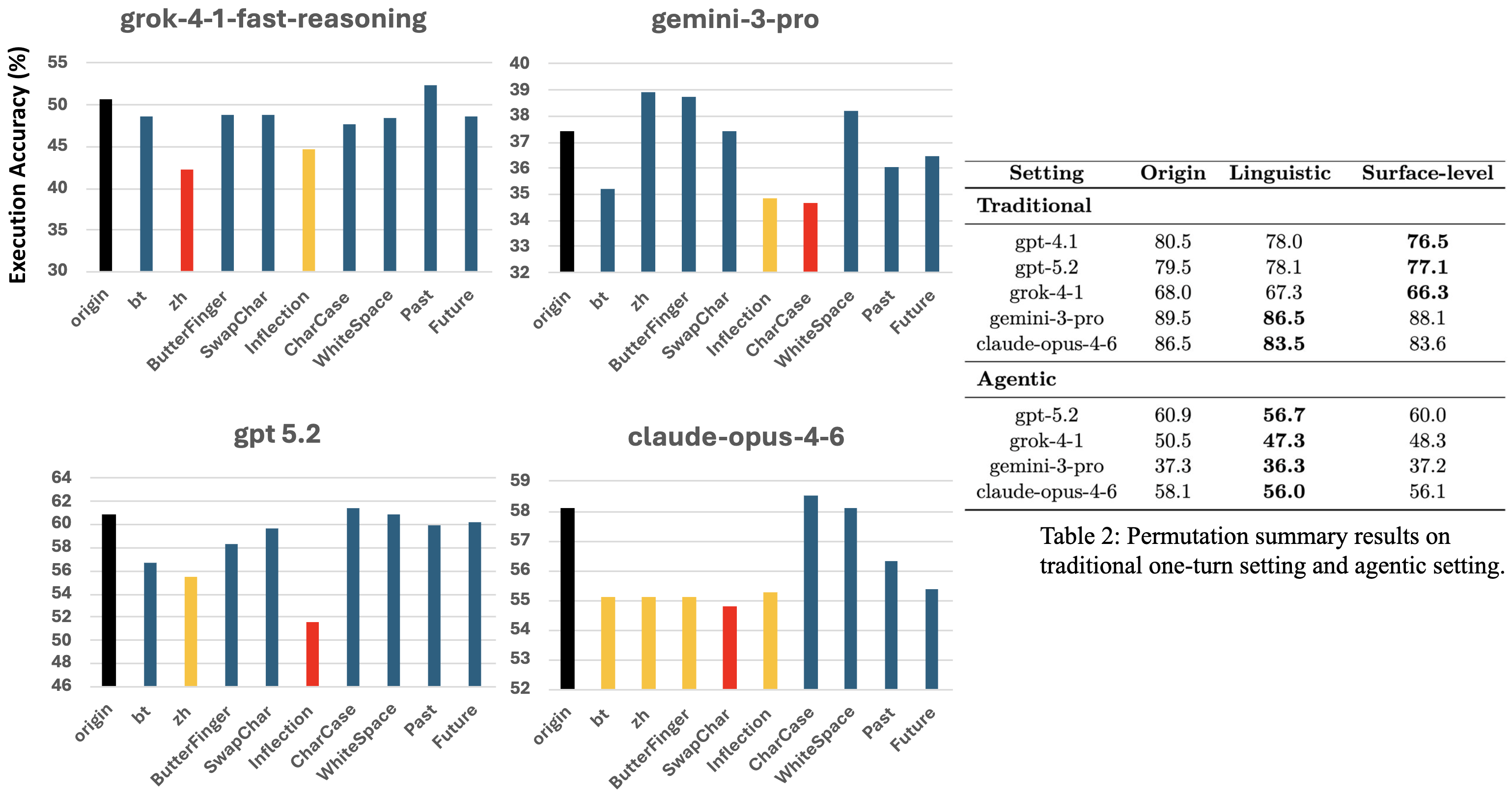}
  \caption{Results for \textbf{R-NL2SQL-Agentic} (adapted from Spider 2~\citep{lei2024spider}). Using the Spider-Agent framework~\citep{lei2024spider} (following ReAct~\citep{yao2023react})for evaluation, we substitute different LLMs. Earlier LLMs (e.g., GPT-4.1) achieved very low results. Traditional pipelines degrade more under surface-level noises, whereas agentic setups are more challenged by linguistic variations.
   }
  \label{fig:spider2}
\end{figure*}

\paragraph{Multiple Databases.} 

In real-world text-to-SQL systems, multiple types of databases may be involved, such as local databases (e.g., SQLite) and cloud databases (e.g., BigQuery). However, in current benchmarks—such as Spider 2.0~\citep{lei2024spider}—the corresponding database for a given query is specified and remains fixed during evaluation. To enable robustness evaluation of agentic text-to-SQL systems, we introduce additional databases for each query. We explore two strategies for this augmentation: randomly and similarity. Setup details in Appendix~\ref{sec:dbnames}.

\subsection{Benchmarks}

In this work, we utilize two datasets—Spider 1~\citep{yu-etal-2018-spider} and Spider 2~\citep{lei2024spider}—to construct our benchmark, \textbf{R-NL2SQL}. For the traditional text-to-SQL setting, we randomly select 200 samples from the Spider 1 development set and apply nine transformations to the user questions. This portion of the benchmark is referred to as \textbf{R-NL2SQL-Tradition}. For the agentic setting, we use 135 samples from Spider 2.0-lite, with all samples leveraging SQLite databases. The \textbf{R-NL2SQL-Agentic} benchmark includes two versions: (1) ten different transformations applied to the user queries, and (2) database updates involving the addition of extra databases—either randomly or based on high similarity of database names—as described in Section~\ref{sec:transform}. 

\subsection{Text-to-SQL Systems}
Five different state-of-the-art LLMs are used in our experiments: GPT-5.2, GPT-4.1, Claude-opus-4-6, Gemini-3-pro, and grok-4-1-fast-reasoning. We use a temperature of 0 for inference. Average results of 5 runs are report. For the traditional setting, we consider zero-shot scenario. For the agentic setting, we use the Spider-Agent~\citep{lei2024spider} (following ReAct~\citep{yao2023react}), which allows for multi-turn interaction with the database via command-line interfaces until the final answer is obtained. Following prior work, we use \textbf{Execution Accuracy(EX)} as our metric.

%The execution accuracy, which compares the execution output of the predicted SQL query with that of the ground truth SQL query on database instances. 
%\section{Quality of Perturbation Samples}

\section{Results}

\subsection{Traditional Settings}

Figure~\ref{fig:spider1} presents robustness evaluation results for five different LLMs on the tradition setting, under various perturbations. LLMs generally maintain strong performance under several perturbations. Across all models, performance drops the most under the ButterFinger (random typos) and back translation, Chinese translation perturbations, indicating particular vulnerability to these types of noise.

% \begin{table}[ht]
%  \centering
%   \scalebox{0.6}{
%   \begin{tabular}{cccc}
%     \hline
%     \textbf{Setting} & \textbf{Origin} & \textbf{Linguistic } & \textbf{Surface-level}  \\
%     \toprule
%     \multicolumn{3}{l}{\textbf{Traditional}} \\
%     \midrule
%     gpt-4.1 & 80.5 & 78.0 & \textbf{76.5} \\
%     gpt-5.2 & 79.5 & 78.1 & \textbf{77.1} \\
%     grok-4-1 & 68.0  & 67.3 & \textbf{66.3} \\
%     gemini-3-pro & 89.5 & \textbf{86.5} & 88.1 \\
%     claude-opus-4-6 & 86.5 & \textbf{83.5} & 83.6 \\
%     \midrule
%     \multicolumn{3}{l}{\textbf{Agentic}} \\
%     \midrule
%     gpt-5.2 & 60.9 & \textbf{56.7} & 60.0 \\
%     grok-4-1 & 50.5  & \textbf{47.3} & 48.3 \\
%     gemini-3-pro & 37.3 & \textbf{36.3} & 37.2 \\
%     claude-opus-4-6 & 58.1 & \textbf{56.0} & 56.1 \\
%     \bottomrule
%   \end{tabular}
%   }
%   \caption{Permutation summary results on traditional one-turn setting and agentic setting.}\label{tab:difference}
%   \end{table}

\subsection{Agentic Settings}

\paragraph{Permutation Results.} Figure~\ref{fig:spider2}~\footnote{Since the GPT-4.1 results for the agentic setting are low (12–15), we have decided to skip the comprehensive perturbation experiments.} displays robustness evaluation results on the R-NL2SQL-Agentic benchmark. Overall, the models maintain relatively stable performance under formatting and grammatical variations such as CharCase, Whitespace. However, \textbf{linguistic variations lead to noticeable performance degradation}. \textbf{Perturbations such as zh, bt, and infection changes cause larger drops across multiple models}, suggesting that NL2SQL systems remain sensitive to lexical noise.

\begin{figure}[h]
  \centering
  \includegraphics[width=0.45\textwidth]{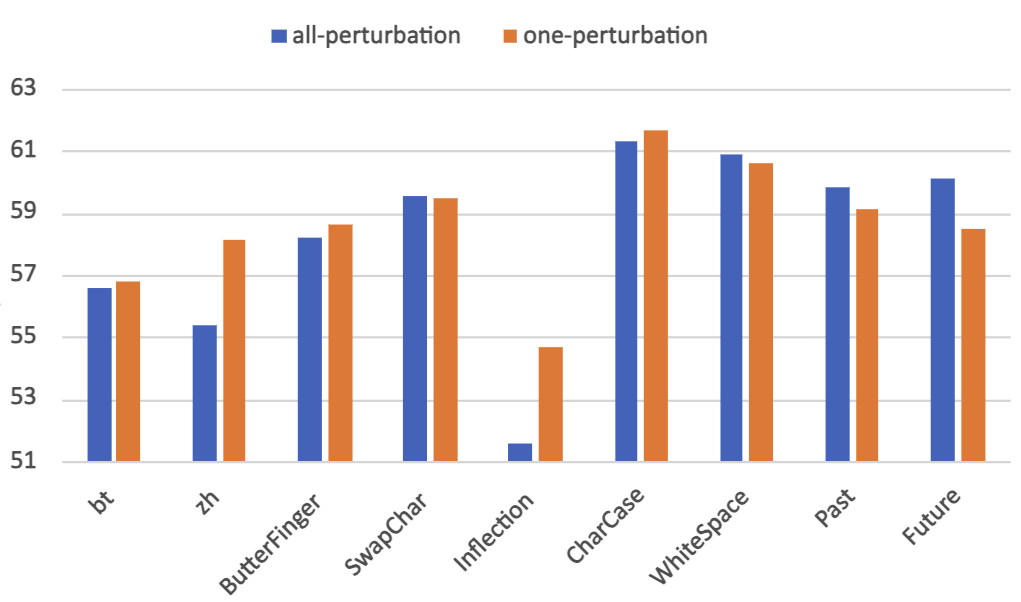}
  \caption{Execution accuracy of ``one-perturbation'' and ``all-perturbation'' in the agentic setting, confirming that linguistic variations larger performance differences. }
  \label{fig:perbs}
\end{figure}

\paragraph{Surface Noise vs. Linguistic
Variation.}
As shown in Table 2, surface-level noise leads to larger performance drops in traditional pipelines, whereas linguistic variations pose greater challenges in agentic settings.
To further investigate and confirm that surface-level noise is less harmful than linguistic variation, we introduce an additional setting, ``one-perturbation'', in which only one randomly selected sentence is perturbed for each query transformation. We compare this setting against ``all-perturbation'', where all sentences are perturbed. Figure~\ref{fig:perbs} shows that linguistic variations (zh, inflection, past, future) cause substantially larger performance drops. Overall, \textbf{linguistic variation is more challenging, and reducing its proportion tends to improve performance.}

\begin{figure}[ht]
  \centering
  \includegraphics[width=0.45\textwidth]{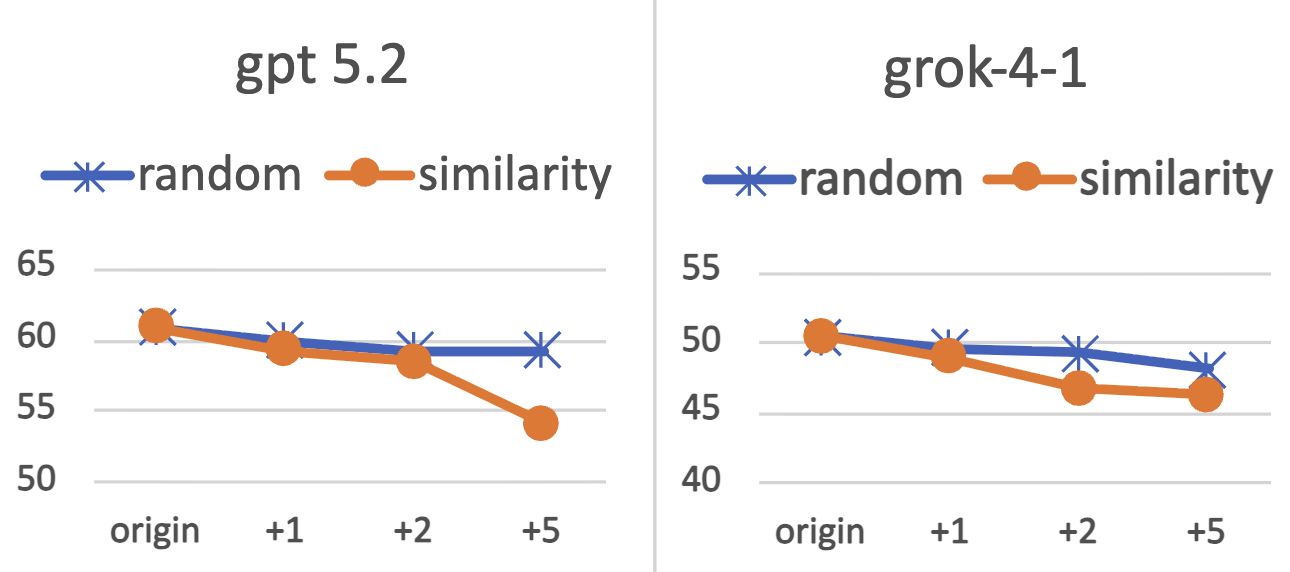}
  \caption{Execution accuracy when extra databases are added in the agentic setting. }
  \label{fig:db}
\end{figure}

\paragraph{Extra DB results.} As shown in Figure~\ref{fig:db}, adding unrelated databases reduces execution accuracy. For both models, similarity-based retrieval/selection is more adversely affected by database clutter than by random selection, indicating another noise-related limitation of the agent framework.

\section{Conclusion}
In this work, we introduce a robustness evaluation benchmark containing approximately ten types of perturbations and conduct evaluations under both traditional and agentic settings. We study two types of noise: (1) surface-level noise (e.g., character-level corruption) and (2) linguistic variation that preserves semantics while changing lexical or syntactic form. We find that surface-level noise causes larger performance drops in traditional pipelines, whereas linguistic variation poses greater challenges in agentic settings.

%\subsection{Appendices}

%Use \verb|\appendix| before any appendix section to switch the section numbering over to letters. See Appendix~\ref{sec:appendix} for an example.

\section*{Limitations}

%In this work, we study two types of noise: (1) surface-level noise (e.g., character-level corruption) and (2) linguistic variation that preserves semantics while changing lexical or syntactic form. During the evaluation, we find the valiance of API call (the output is not deterministic even temperature is 0, and seed is set). In order to get a more reliable evaluation, we report the average results of 5 runs. More run could be beneficial, however, we set to 5  because of the high cost. In addition, in our experiments, there are only about 150-200 samples for each perturbation. In this work, we suggest that linguistic variation presents greater challenges in agentic settings. It would be interesting to explore whether some argentic setting can have lower drop on this setting.  

In this work, we study two types of noise: (1) surface-level noise (e.g., character-level corruption) and (2) linguistic variation that preserves semantics while altering lexical or syntactic form. During evaluation, we observed variance in API calls (outputs were non-deterministic even with temperature set to 0 and a fixed seed). To obtain more reliable results, we report the average over five runs. While additional runs could further improve stability, we limit to five due to cost. Each perturbation condition includes approximately 130–200 samples. Our results suggest that linguistic variation poses greater challenges in agentic settings. Future work could explore agentic configurations that mitigate this drop.

Another limitation concerns the comparison between the two noise types. An ideal analysis would provide an apples-to-apples comparison—reporting, for both datasets, performance under both the traditional and agentic settings. In practice, however, the two benchmarks rely on different pipelines. As shown in Spider 2~\citep{lei2024spider}, Spider 2 poses substantial challenges for traditional text-to-SQL methods, while Spider 1 is comparatively easy and thus not a good fit for evaluating agentic systems. To further validate the effects of the two noise types, we therefore consider two permutation ratios (``all-perturbation`` and ``one-perturbation``). Overall,  as shown in Figure~\ref{fig:perbs}, linguistic variation is more challenging: reducing the proportion of linguistic-variation perturbations tends to yield better performance.

\bibliography{custom}

\begin{thebibliography}{20}
\providecommand{\natexlab}[1]{#1}

\bibitem[{Anthropic(2026)}]{anthropic2026claudeopus46}
Anthropic. 2026.
\newblock \href {https://www.anthropic.com/news/claude-opus-4-6} {Introducing {Claude} {Opus} 4.6}.
\newblock Anthropic News.
\newblock Accessed: 2026-03-16.

\bibitem[{Chang et~al.(2023)Chang, Wang, Dong, Pan, Zhu, Li, Lan, Zhang, Jiang, Lilien et~al.}]{chang2023dr}
Shuaichen Chang, Jun Wang, Mingwen Dong, Lin Pan, Henghui Zhu, Alexander~Hanbo Li, Wuwei Lan, Sheng Zhang, Jiarong Jiang, Joseph Lilien, and 1 others. 2023.
\newblock Dr. spider: A diagnostic evaluation benchmark towards text-to-sql robustness.
\newblock \emph{arXiv preprint arXiv:2301.08881}.

\bibitem[{DeepMind(2025)}]{gemini3pro}
Google DeepMind. 2025.
\newblock Gemini 3 pro: Model card and product page.
\newblock \url{https://ai.google.dev/gemini-api/docs/models#gemini-3-pro}.

\bibitem[{Deng et~al.(2021)Deng, Awadallah, Meek, Polozov, Sun, and Richardson}]{deng-etal-2021-structure}
Xiang Deng, Ahmed~Hassan Awadallah, Christopher Meek, Oleksandr Polozov, Huan Sun, and Matthew Richardson. 2021.
\newblock \href {https://doi.org/10.18653/v1/2021.naacl-main.105} {Structure-grounded pretraining for text-to-{SQL}}.
\newblock In \emph{Proceedings of the 2021 Conference of the North American Chapter of the Association for Computational Linguistics: Human Language Technologies}, pages 1337--1350, Online. Association for Computational Linguistics.

\bibitem[{Dhole et~al.(2022)Dhole, Gangal, Gehrmann, Gupta, Li, Mahamood, Mahendiran, Mille, Shrivastava, Tan, Wu, Sohl-Dickstein, Choi, Hovy, Dusek, Ruder, Anand, Aneja, Banjade, Barthe, Behnke, Berlot-Attwell, Boyle, Brun, Cabezudo, Cahyawijaya, Chapuis, Che, Choudhary, Clauss, Colombo, Cornell, Dagan, Das, Dixit, Dopierre, Dray, Dubey, Ekeinhor, Giovanni, Goyal, Gupta, Gupta, Hamla, Han, Harel-Canada, Honore, Jindal, Joniak, Kleyko, Kovatchev, Krishna, Kumar, Langer, Lee, Levinson, Liang, Liang, Liu, Lukyanenko, Marivate, de~Melo, Meoni, Meyer, Mir, Moosavi, Muennighoff, Mun, Murray, Namysl, Obedkova, Oli, Pasricha, Pfister, Plant, Prabhu, Pais, Qin, Raji, Rajpoot, Raunak, Rinberg, Roberts, Rodriguez, Roux, S., Sai, Schmidt, Scialom, Sefara, Shamsi, Shen, Shi, Shi, Shvets, Siegel, Sileo, Simon, Singh, Sitelew, Soni, Sorensen, Soto, Srivastava, Srivatsa, Sun, T, Tabassum, Tan, Teehan, Tiwari, Tolkiehn, Wang, Wang, Wang, Wang, Wei, Wilie, Winata, Wu, Wydmański, Xie, Yaseen, Yee, Zhang, and
  Zhang}]{dhole2022nlaugmenter}
Kaustubh~D. Dhole, Varun Gangal, Sebastian Gehrmann, Aadesh Gupta, Zhenhao Li, Saad Mahamood, Abinaya Mahendiran, Simon Mille, Ashish Shrivastava, Samson Tan, Tongshuang Wu, Jascha Sohl-Dickstein, Jinho~D. Choi, Eduard Hovy, Ondrej Dusek, Sebastian Ruder, Sajant Anand, Nagender Aneja, Rabin Banjade, and 107 others. 2022.
\newblock \href {https://arxiv.org/abs/2112.02721} {Nl-augmenter: A framework for task-sensitive natural language augmentation}.
\newblock \emph{Preprint}, arXiv:2112.02721.

\bibitem[{Dong et~al.(2023)Dong, Zhang, Ge, Mao, Gao, lu~Chen, Lin, and Lou}]{dong2023c3zeroshottexttosqlchatgpt}
Xuemei Dong, Chao Zhang, Yuhang Ge, Yuren Mao, Yunjun Gao, lu~Chen, Jinshu Lin, and Dongfang Lou. 2023.
\newblock \href {https://arxiv.org/abs/2307.07306} {C3: Zero-shot text-to-sql with chatgpt}.
\newblock \emph{Preprint}, arXiv:2307.07306.

\bibitem[{Gan et~al.(2021)Gan, Chen, Huang, Purver, Woodward, Xie, and Huang}]{gan-etal-2021-towards}
Yujian Gan, Xinyun Chen, Qiuping Huang, Matthew Purver, John~R. Woodward, Jinxia Xie, and Pengsheng Huang. 2021.
\newblock \href {https://doi.org/10.18653/v1/2021.acl-long.195} {Towards robustness of text-to-{SQL} models against synonym substitution}.
\newblock In \emph{Proceedings of the 59th Annual Meeting of the Association for Computational Linguistics and the 11th International Joint Conference on Natural Language Processing (Volume 1: Long Papers)}, pages 2505--2515, Online. Association for Computational Linguistics.

\bibitem[{Gao et~al.(2023)Gao, Wang, Li, Sun, Qian, Ding, and Zhou}]{dail_sql}
Dawei Gao, Haibin Wang, Yaliang Li, Xiuyu Sun, Yichen Qian, Bolin Ding, and Jingren Zhou. 2023.
\newblock Text-to-sql empowered by large language models: A benchmark evaluation.
\newblock \emph{CoRR}, abs/2308.15363.

\bibitem[{Lei et~al.(2024)Lei, Chen, Ye, Cao, Shin, Su, Suo, Gao, Hu, Yin et~al.}]{lei2024spider}
Fangyu Lei, Jixuan Chen, Yuxiao Ye, Ruisheng Cao, Dongchan Shin, Hongjin Su, Zhaoqing Suo, Hongcheng Gao, Wenjing Hu, Pengcheng Yin, and 1 others. 2024.
\newblock Spider 2.0: Evaluating language models on real-world enterprise text-to-sql workflows.
\newblock \emph{arXiv preprint arXiv:2411.07763}.

\bibitem[{Liu et~al.(2025)Liu, Tan, Zhong, Xie, Zhao, Wang, Hu, and Li}]{liu-etal-2025-solid}
Geling Liu, Yunzhi Tan, Ruichao Zhong, Yuanzhen Xie, Lingchen Zhao, Qian Wang, Bo~Hu, and Zang Li. 2025.
\newblock \href {https://aclanthology.org/2025.coling-main.654/} {Solid-{SQL}: Enhanced schema-linking based in-context learning for robust text-to-{SQL}}.
\newblock In \emph{Proceedings of the 31st International Conference on Computational Linguistics}, pages 9793--9803, Abu Dhabi, UAE. Association for Computational Linguistics.

\bibitem[{OpenAI(2025{\natexlab{a}})}]{openai2025gpt41}
OpenAI. 2025{\natexlab{a}}.
\newblock \href {https://openai.com/index/gpt-4-1/} {{Introducing GPT‑4.1 in the API }}.
\newblock Accessed: [Insert Date Here].

\bibitem[{OpenAI(2025{\natexlab{b}})}]{openai2025gpt52}
OpenAI. 2025{\natexlab{b}}.
\newblock \href {https://openai.com/index/introducing-gpt-5-2/} {{Introducing GPT‑5.2}}.

\bibitem[{Pourreza and Rafiei(2023)}]{pourreza2023dinsql}
Mohammadreza Pourreza and Davood Rafiei. 2023.
\newblock \href {https://openreview.net/forum?id=p53QDxSIc5} {{DIN}-{SQL}: Decomposed in-context learning of text-to-{SQL} with self-correction}.
\newblock In \emph{Thirty-seventh Conference on Neural Information Processing Systems}.

\bibitem[{Safarzadeh et~al.(2025)Safarzadeh, Oroojlooy, and Roth}]{safarzadeh-etal-2025-evaluating}
Mohammadtaher Safarzadeh, Afshin Oroojlooy, and Dan Roth. 2025.
\newblock \href {https://doi.org/10.18653/v1/2025.findings-emnlp.1031} {Evaluating {NL}2{SQL} via {SQL}2{NL}}.
\newblock In \emph{Findings of the Association for Computational Linguistics: EMNLP 2025}, pages 18954--18968, Suzhou, China. Association for Computational Linguistics.

\bibitem[{Saparina and Lapata(2024)}]{saparina-lapata-2024-improving}
Irina Saparina and Mirella Lapata. 2024.
\newblock \href {https://doi.org/10.18653/v1/2024.eacl-long.71} {Improving generalization in semantic parsing by increasing natural language variation}.
\newblock In \emph{Proceedings of the 18th Conference of the European Chapter of the Association for Computational Linguistics (Volume 1: Long Papers)}, pages 1178--1193, St. Julian{'}s, Malta. Association for Computational Linguistics.

\bibitem[{Tan et~al.(2020)Tan, Joty, Kan, and Socher}]{tan-etal-2020-morphin}
Samson Tan, Shafiq Joty, Min-Yen Kan, and Richard Socher. 2020.
\newblock \href {https://doi.org/10.18653/v1/2020.acl-main.263} {It{'}s morphin' time! {C}ombating linguistic discrimination with inflectional perturbations}.
\newblock In \emph{Proceedings of the 58th Annual Meeting of the Association for Computational Linguistics}, pages 2920--2935, Online. Association for Computational Linguistics.

\bibitem[{xAI(2025)}]{xai_grok41_2025}
xAI. 2025.
\newblock \href {https://x.ai/news/grok-4-1} {Grok 4.1 fast and agent tools api}.

\bibitem[{Yao et~al.(2023)Yao, Zhao, Yu, Du, Shafran, Narasimhan, and Cao}]{yao2023react}
Shunyu Yao, Jeffrey Zhao, Dian Yu, Nan Du, Izhak Shafran, Karthik Narasimhan, and Yuan Cao. 2023.
\newblock {ReAct}: Synergizing reasoning and acting in language models.
\newblock In \emph{International Conference on Learning Representations (ICLR)}.

\bibitem[{Yu et~al.(2018)Yu, Zhang, Yang, Yasunaga, Wang, Li, Ma, Li, Yao, Roman, Zhang, and Radev}]{yu-etal-2018-spider}
Tao Yu, Rui Zhang, Kai Yang, Michihiro Yasunaga, Dongxu Wang, Zifan Li, James Ma, Irene Li, Qingning Yao, Shanelle Roman, Zilin Zhang, and Dragomir Radev. 2018.
\newblock \href {https://doi.org/10.18653/v1/D18-1425} {{S}pider: A large-scale human-labeled dataset for complex and cross-domain semantic parsing and text-to-{SQL} task}.
\newblock In \emph{Proceedings of the 2018 Conference on Empirical Methods in Natural Language Processing}, pages 3911--3921, Brussels, Belgium. Association for Computational Linguistics.

\bibitem[{Zhang et~al.(2025)Zhang, Qian, Sahai, Tian, Garg, Sun, and Li}]{Evoschema}
Tianshu Zhang, Kun Qian, Siddhartha Sahai, Yuan Tian, Shaddy Garg, Huan Sun, and Yunyao Li. 2025.
\newblock \href {https://doi.org/10.14778/3748191.3748222} {Evoschema: Towards text-to-sql robustness against schema evolution}.
\newblock \emph{Proc. VLDB Endow.}, 18(10):3655–3668.

\end{thebibliography}

\appendix

In this work, AI assistants (ChatGPT) is used for polish the writing only. 

\section{LLMs Used in Our Experiments}
\begin{itemize}
    \item GPT-4.1~\citep{openai2025gpt41}
    \item GPT-5.2~\citep{openai2025gpt52}
    \item Claude-opus-4-6~\citep{anthropic2026claudeopus46}
    \item Gemini-3-pro~\citep{gemini3pro}
    \item grok-4-1-fast-reasoning~\citep{xai_grok41_2025}.
\end{itemize}

\section{Transformations on Queries}
\label{sec:appendixA}

Following are the detailed description for each transformation on a query:

\paragraph{Chinese (zh).} This transformation translates a given English query into Chinese. It changes the language of the query which preserve the content. It is one example to evaluate the multilingual generalization capability.  In our experiments, we use GPT-5.2 as the English-Chinese Translator. 

\paragraph{BackTranslation (bt).} This transformation translated a given English query into Chinese and back to English. This transformation acts like a paraphraser. Usually, multiple variation can be created. In the implementation, GPT-5.2 is used. Check Appendix~\ref{sec:prompt} for the prompt template. 

\paragraph{EnglishInflection.} This transformation randomly selects words in the query and change them to a random inflection variance. This can be from plural to singular (or vice versa) for ouns and tense changes for verbs. Perturbation is constrained to be
the same Part of Speech (POS) tag. This is adapted from the Morpheus~\citep{tan-etal-2020-morphin} adversarial attack.

\paragraph{Past.} This is a deterministic transformation that converts sentences in the query to past tense.

\paragraph{Future.} This is a deterministic transformation that converts sentences in the query to future tense.

\paragraph{ButterFinger} This transformation adds noise  erupting from keyboard typos making common spelling errors. Few letters picked at random are replaced with letters which are at keyboard positions near the source letter. The implementation is borrowed from NL-Augmenter~\citep{dhole2022nlaugmenter}.

\paragraph{ChangeCharCase.} This transformation randomly changes the selected characters to upper case in the query. 

\paragraph{SwapChar.} This transformation randomly
selects pairs of adjacent characters in the query
and swap them. To ensure naturalness, we set the
probability as 0.05 for making the swap.

\paragraph{WhiteSpace.} This transformation inserts or
deletes a single white space at randomly selected locations in the docstring. This represents a common
type of typos by humans. Following NL-Augmenter~\citep{dhole2022nlaugmenter},
we use probability 0.1 for adding whitespaces and
0.05 for removing whitespaces.

\subsection{Template for Chinese translation and BackTranslation}
\label{sec:prompt}

Give an English query \texttt{\{q\}}, the following template is used to generate a query: 

\begin{quote}
    \texttt{Task: Translate the provided sentence into Chinese and then back into English. Return the result strictly as a JSON object. \\\\Instructions:\\\\
    Translate the English sentence into natural Mandarin Chinese.\\\\Translate that Chinese version back into English, capturing the literal meaning of the Chinese phrasing.\\\\ Provide a "delta\_analysis" field identifying any nuances lost or gained (e.g., changes in formality or specific idioms). \\\\Output Schema:\\ \{ \\
        "original\_en": "string",\\
        "translated\_zh": "string",\\
        "back\_translated\_en": "string",\\
        "delta\_analysis": "string"\\\}\\\\Sentence to translate: "\{q\}"\\
    }
\end{quote}

\section{Multiple databases}
\label{sec:dbnames}
%Below are the five most semantically similar databases. In the agentic setting,additional databases are included in the experimental environment.
\subsection{Setup}

Following are the two different ways to add extra databases in the agentic setting. 
\begin{itemize}
\item \textbf{Random:} In this setting, in addition to the corresponding database, we randomly add 1, 2, or 5 extra databases during evaluation for each user query.
\item \textbf{High Similarity:} In this setting, for each database, we select up to 5 additional databases that are most similar based on the semantic similarity of their database names. Then for each query, the additional 1, 2, or 5 databases are selected among these similar databases of the corresponding database. Check more related details in Appendix~\ref{sec:dbnames}.  
\end{itemize}

\subsection{How to select the extra databases?}

The five most semantically similar databases for each ground-truth database are shown below. Under the agentic setting, additional databases are added to the experimental environment. Table~\ref{tab:dbnames} illustrations of the five most similar databases for each database.

\begin{table*}[h]
\footnotesize
\centering

\setlength{\tabcolsep}{1pt}
\scalebox{0.8}{
\begin{tabular}{|c|c|} \toprule
db & \multicolumn{1}{|c|}{5 database names that are most similar}                                                                             \\ \midrule

bank\_sales\_trading & electronic\_sales, E\_commerce, Brazilian\_E\_Commerce, northwind, AdventureWorks \\
Pagila & sqlite-sakila, chinook, Db-IMDB, imdb\_movies, northwind \\
electronic\_sales & E\_commerce, Brazilian\_E\_Commerce, bank\_sales\_trading, northwind, AdventureWorks \\
E\_commerce & Brazilian\_E\_Commerce, electronic\_sales, bank\_sales\_trading, northwind, AdventureWorks \\
Brazilian\_E\_Commerce & E\_commerce, electronic\_sales, bank\_sales\_trading, northwind, AdventureWorks \\
complex\_oracle & oracle\_sql, modern\_data, log, AdventureWorks, northwind \\
California\_Traffic\_Collision & city\_legislation, f1, Airlines, delivery\_center, EU\_soccer \\
EU\_soccer & IPL, Baseball, f1, WWE, BowlingLeague \\
stacking & modern\_data, log, complex\_oracle, delivery\_center, school\_scheduling \\
imdb\_movies & Db-IMDB, EntertainmentAgency, music, chinook, Pagila \\
Db-IMDB & imdb\_movies, Pagila, chinook, EntertainmentAgency, music \\
WWE & EU\_soccer, IPL, Baseball, f1, EntertainmentAgency \\
Baseball & EU\_soccer, IPL, f1, WWE, BowlingLeague \\
northwind & AdventureWorks, E\_commerce, electronic\_sales, Brazilian\_E\_Commerce, bank\_sales\_trading \\
f1 & EU\_soccer, IPL, Baseball, WWE, Airlines \\
EntertainmentAgency & imdb\_movies, Db-IMDB, music, WWE, EU\_soccer \\
education\_business & school\_scheduling, northwind, AdventureWorks, modern\_data, stacking \\
music & chinook, imdb\_movies, Db-IMDB, EntertainmentAgency, Pagila \\
sqlite-sakila & Pagila, chinook, Db-IMDB, imdb\_movies, northwind \\
chinook & sqlite-sakila, Pagila, music, Db-IMDB, imdb\_movies \\
modern\_data & stacking, complex\_oracle, log, AdventureWorks, northwind \\
log & modern\_data, stacking, complex\_oracle, oracle\_sql, delivery\_center \\
oracle\_sql & complex\_oracle, modern\_data, log, AdventureWorks, northwind \\
school\_scheduling & education\_business, BowlingLeague, northwind, AdventureWorks, modern\_data \\
IPL & EU\_soccer, Baseball, f1, WWE, BowlingLeague \\
city\_legislation & California\_Traffic\_Collision, modern\_data, log, education\_business, school\_scheduling \\
AdventureWorks & northwind, E\_commerce, electronic\_sales, Brazilian\_E\_Commerce, bank\_sales\_trading \\
delivery\_center & Airlines, stacking, modern\_data, AdventureWorks, log \\
BowlingLeague & Baseball, EU\_soccer, IPL, WWE, school\_scheduling \\
Airlines & delivery\_center, f1, California\_Traffic\_Collision, AdventureWorks, modern\_data 

\\ \bottomrule
\end{tabular}
}
\caption{Illustrations of the five most similar databases for each database. The similarity are based on judged by LLMs.}
\label{tab:dbnames}

\end{table*}

\section{Text-to-SQL Systems Setup}
Five different state-of-the-art LLMs are used in our experiments: GPT-5.2, GPT-4.1, Claude-opus-4-6, Gemini-3-pro, and grok-4-1-fast-reasoning. We use a temperature of 0 for inference.

For the traditional setting, we consider zero-shot scenario. For the agentic setting, we use the Spider-Agent, which allows for multi-turn interaction with the database via command-line interfaces until the final answer is obtained. 

\paragraph{Metric.} Following prior work, we use execution accuracy(EX) as our metric. The execution accuracy, which compares the execution output of the predicted SQL query with that of the ground truth SQL query on database instances.

\section{Related Work}

Several prior studies have explored the robustness and generalization capabilities of Natural Language to SQL (NL2SQL) systems~\citep{gan-etal-2021-towards, deng-etal-2021-structure,saparina-lapata-2024-improving,Evoschema,liu-etal-2025-solid}. Dr.Spider~\citep{chang2023dr} extends the Spider benchmark by introducing multiple perturbations, such as linguistic variations and noise injection, to systematically evaluate model robustness under distribution shifts. Another line of work focuses on linguistic robustness. \citet{safarzadeh-etal-2025-evaluating} leverage SQL-to-NL generation to produce schema-aligned paraphrases of queries, enabling evaluation of NL2SQL models under semantically equivalent but linguistically diverse inputs. These studies highlight that previous models remain sensitive to input perturbations in traditional settings. In our work, we conduct empirical evaluations of several state-of-the-art models under both traditional and agentic settings.

\section{Results of An Eariler LLM on Agentic Setting}

\begin{figure}[h]
  \centering
  \includegraphics[width=0.45\textwidth]{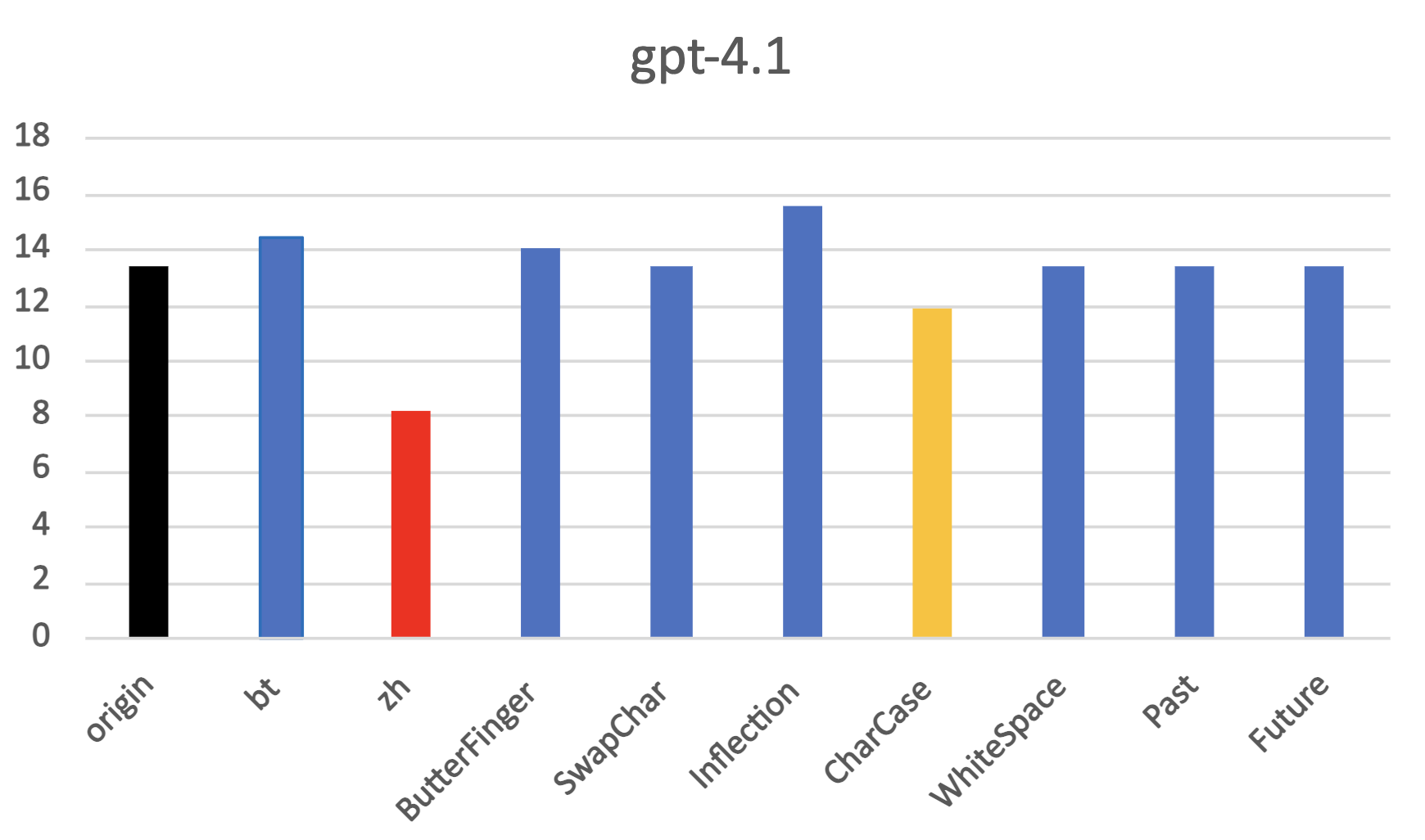}
  \caption{Execution accuracy of GPT-4.1 in the agentic setting. The low accuracy indicates that earlier LLMs have much weaker agentic capabilities.}
  \label{fig:gpt4}
\end{figure}

\end{document}